\documentclass[conference]{IEEEtran}
\IEEEoverridecommandlockouts
\usepackage{cite}
\usepackage{amsmath,amssymb,amsfonts}
\usepackage{algorithmic}
\usepackage{graphicx}
\usepackage{textcomp}
\usepackage{xcolor}
\usepackage{mathrsfs} 
\usepackage{amsmath}
\usepackage{amsmath,amsfonts}
\usepackage{algorithmic}
\usepackage{array}
\usepackage[caption=false,font=normalsize,labelfont=sf,textfont=sf]{subfig}
\usepackage{textcomp}
\usepackage{stfloats}
\usepackage{url}
\usepackage{pifont}% http://ctan.org/pkg/pifont
\usepackage{verbatim}
\usepackage{graphicx}
\usepackage{amssymb}
\usepackage{amsmath,amssymb,amsfonts,bm}
\usepackage{flushend}
\usepackage{mathrsfs}
\usepackage{graphicx}
\usepackage{float}
\floatstyle{plaintop}
\restylefloat{table}
\usepackage{textcomp}
\usepackage{physics}
\usepackage{tablefootnote}
\usepackage{epstopdf}    % Para poder insertar %figuras .eps al compilar con PDFLATEX
\usepackage{flushend}       
\usepackage{amsthm}

\usepackage{array}
\usepackage{mdwtab}
\usepackage{eqparbox}
\usepackage{amsthm}
\usepackage{hyperref}
\usepackage[export]{adjustbox}
\usepackage{wrapfig}
\usepackage{lipsum}
\usepackage{graphicx}
\usepackage{adjustbox}
\usepackage[normalem]{ulem}
\usepackage[numbers]{natbib}
\useunder{\uline}{\ul}{}
\def\BibTeX{{\rm B\kern-.05em{\sc i\kern-.025em b}\kern-.08em
    T\kern-.1667em\lower.7ex\hbox{E}\kern-.125emX}}
\begin{document}

\title{R$^2$VFL: A Robust Random Vector Functional Link Network with Huber-Weighted Framework\\
\author{
\IEEEauthorblockN{Anuradha Kumari}
\IEEEauthorblockA{
\textit{Department of Mathematics} \\
\textit{Indian Institute of Technology Indore}\\
phd2101141007@iiti.ac.in}
\and
\IEEEauthorblockN{Mushir Akhtar}
\IEEEauthorblockA{
\textit{Department of Mathematics} \\
\textit{Indian Institute of Technology Indore}\\
phd2101241004@iiti.ac.in}
\and
\IEEEauthorblockN{P. N. Suganthan}
\IEEEauthorblockA{
\textit{KINDI Center for Computing Research} \\
\textit{College of Engineering}\\
\textit{ Qatar University, Doha 2713}\\
p.n.suganthan@gmail.com}
\and
\IEEEauthorblockN{M. Tanveer\textsuperscript{*}\thanks{\textsuperscript{*}Corresponding author}}
\IEEEauthorblockA{
\textit{Department of Mathematics} \\
\textit{Indian Institute of Technology Indore}\\
mtanveer@iiti.ac.in}
}
}

\maketitle

\begin{abstract}
The random vector functional link (RVFL) neural network has shown significant potential in overcoming the constraints of traditional artificial neural networks, such as excessive computation time and suboptimal solutions. However, RVFL faces challenges when dealing with noise and outliers, as it assumes all data samples contribute equally. To address this issue, we propose a novel robust framework, R$^2$VFL, RVFL with Huber weighting function and class probability, which enhances the model’s robustness and adaptability by effectively mitigating the impact of noise and outliers in the training data. The Huber weighting function reduces the influence of outliers, while the class probability mechanism assigns less weight to noisy data points, resulting in a more resilient model. 
We explore two distinct approaches for calculating class centers within the R$^2$VFL framework: the simple average of all data points in each class and the median of each feature, the later providing a robust alternative by minimizing the effect of extreme values. These approaches give rise to two novel variants of the model—R$^2$VFL-A and R$^2$VFL-M. 
We extensively evaluate the proposed models on 47 UCI datasets, encompassing both binary and multiclass datasets, and conduct rigorous statistical testing, which confirms the superiority of the proposed models. Notably, the models also demonstrate exceptional performance in classifying EEG signals, highlighting their practical applicability in real-world biomedical domain. %This work is a critical step forward in making RVFL more robust and capable of handling real-world noisy and outlier-prone data, setting a new standard for performance and adaptability in machine learning applications.

% The plane-based classifiers, support vector machine (SVM) and twin support vector machine (TWSVM), are susceptible to the negative impact of noise, outliers, and class imbalance learning (CIL). The concept of intuitionistic fuzzy (IF) addresses the impact of noise and outliers present in the data. However, IF is sensitive to the threshold distance value and calculates the distance of each data point to other points of its class which is computatinally intensive. Moreover, IF assigns the same score value to the data points that have distinct distances from the center of the other class and identical membership values. To address the aforementioned limitations, in this paper, we propose dual center-based intuitionistic fuzzy plane-based classifiers, specifically dual center-based intuitionistic fuzzy support vector machine (DC-IFSVM) and dual center-based intuitionistic fuzzy least square twin support vector machine (DC-IFLSTSVM). The proposed models, DC-IFSVM and DC-IFLSTSVM handle the CIL by associating class-specific terms in the weighting function. The optimization problem of DC-IFLSTSVM is solved using the conjugate gradient (CG) method. We conducted extensive numerical experiments of the proposed DC-IFSVM, DC-IFLSTSVM and baseline models on 54 UCI and KEEL datasets, which resulted in the superiority of the proposed DC-IFLSTSVM. Furthermore, experiments on the BreakHis dataset, focused on classifying benign tumors from malignant tumors, showcased the remarkable performance of the proposed DC-IFLSTSVM.  
\end{abstract}


\section{Introduction}
\IEEEPARstart{A}{mong} the several machine learning techniques, artificial neural networks' (ANNs') capacity to approximate nonlinear mappings is the main factor contributing to their enormous success across a wide range of fields \cite{abiodun2018state}. ANNs process, analyse and transmit data to generate predictions or decisions. ANNs have shown success in several domains, including stock market forecasting \cite{dase2010application}, brain age prediction \cite{tanveer2023deep}, and more. \par %rainfall forecasting \cite{luk2001application}, clinical medicine \cite{baxt1995application}
One of the most widely used techniques for training ANNs is the gradient descent (GD) method, which operates through an iterative process. This method optimizes the model's weights and biases by backpropagating the error—the difference between the actual and predicted outputs. Despite its popularity, the GD-based approach faces several inherent challenges. It is often time-consuming, susceptible to converging to local optima rather than the global optimum \cite{gori1992problem}, and highly sensitive to both the initial parameter settings and the choice of learning rate. \par
To overcome the limitations of GD-based neural networks, randomized neural networks (RNNs) \cite{schmidt1992feed, pao1994learning} have been introduced as an alternative. These networks incorporate a degree of randomness into their structure or learning process. Typically, this randomness is applied by using a closed-form solution to determine the parameters of the output layer while keeping certain training parameters fixed \cite{suganthan2018non}.  Due to their inherent randomness, RNNs can learn efficiently, requiring fewer tunable parameters and operating without the need for expensive hardware. Among the most notable RNNs is the random vector functional link (RVFL) neural network \cite{pao1994learning, malik2023random}. In RVFL, the weights and biases of the hidden layer are randomly initialized from an appropriate continuous probability distribution and remain fixed throughout the training process. A defining feature of RVFL is its direct connections between the input and output layers, which distinguish it from other RNNs. These direct linkages act as implicit regularization for the randomization process, significantly enhancing the learning performance of RVFL, as demonstrated in previous studies \cite{zhang2016comprehensive}. The output parameters of RVFL, specifically the direct link weights and the weights connecting the hidden layer to the output layer, are analytically determined using methods such as the pseudo-inverse or least squares. The inclusion of direct linkages in RVFL serves as a form of regularization for the randomization process, which has been shown to significantly enhance the learning performance of the network \cite{zhang2016comprehensive}. \par

Numerous variants of RVFL have been introduced in the literature, to improve the generalization performance. RVFL with privileged information leads to the introduction of RVFL+ and KRVFL+ \cite{zhang2020new},  sparse
autoencoder with $l_1$-norm regularization introduced SP-RVFL \cite{zhang2019unsupervised}. The effectiveness of the network's learning process is heavily influenced by the number of hidden nodes \cite{li2017bayesian}. To address this, \citet{dai2022incremental} introduced incremental RVFL+ (IRVFL+), a constructive algorithm that incrementally adds hidden nodes, allowing the network to approximate the output better. A kernel-based approach, known as the kernel-based exponentially expanded RVFL (KERVFL), is proposed in \cite{chakravorti2020non}. By integrating a kernel function into the RVFL model, KERVFL eliminates the need to determine the optimal number of hidden nodes, offering a more flexible and efficient solution. Some other RVFL variants are as follows: improved parameter learning methodology for RVFL \cite{sun2023improved}, graph embedded RVFL \cite{ganaie2024graph}, angle-based twin RVFL \cite{mishra2024efficient}, and so on. Further, RVFL has wide applications such as Alzheimer's disease diagnosis \cite{sharma2022conv}, quality prediction of blast furnace \cite{zhou2021improved}, fault diagnosis \cite{li2024impact}. \par
Despite these advancements, challenges such as class noise and outliers continue to hinder RVFL's performance. Class noise refers to errors in class labels, where instances are mislabeled, while outliers are data points that deviate significantly from the majority. Both issues introduce inaccuracies and instability, impacting the overall effectiveness of RVFL-based models.
% Class noise refers to errors in class labels where instances are incorrectly labeled, while outliers are data points that deviate significantly from the majority. Both can adversely affect the performance of machine learning models like RVFL by introducing inaccuracies and instability. 
To address these challenges, researchers have developed various RVFL-based models, including intuitionistic fuzzy RVFL (IFRVFL) \cite{malik2022alzheimer}, minimum variance embedded IFRVFL \cite{ahmad2022minimum}, and Bayesian RVFL \cite{scardapane2017bayesian}.

Although the intuitionistic fuzzy scheme \cite{malik2022alzheimer} demonstrates robustness against noise and outliers, it has limitations with certain data distributions. For instance, when most data points cluster near the class center, the intuitionistic fuzzy scheme assigns the highest membership value only to the class center, while other points receive membership values below 1. Similarly, if the majority of data points are concentrated near the class boundary, their membership values are reduced, thereby diminishing their contribution to the training process.

% The Huber weighting function \cite{schuster2011robust}, derived from the widely used Huber loss \cite{balasundaram2020robust}, offers an effective solution to handle outliers in robust regression tasks. 
% It seamlessly combines squared and absolute errors, providing a smooth transition between them, making it both robust to outliers and efficient for smaller errors. 
The Huber weighting function \cite{schuster2011robust}, derived from the widely recognized Huber loss \cite{balasundaram2020robust}, provides an effective mechanism for addressing outliers in robust regression tasks. While the Huber weighting scheme is adept at mitigating the influence of outliers, it may be adversely affected by class noise. To overcome this limitation, we incorporate class probability alongside the Huber weighting function. In this paper, we propose a novel RVFL model, termed R$^2$VFL, which leverages the Huber weighting function to improve robustness against outliers and incorporates class probability to address the effects of noise. The computation of the Huber weights requires determining the class center, which can be achieved through two distinct approaches: by averaging the data points or by using the median of each feature as the class center. This dual approach ensures flexibility and improved performance in various data distributions. The contributions of the paper are as follows:  
\begin{itemize}
    \item We propose novel RVFL model with a Huber weighting scheme and class probability, termed R$^2$VFL, designed to handle noise and outliers in data effectively.
    \item The Huber weighting function depends on accurately determining the class center to ensure robustness against noise and outliers. To achieve this, we present two distinct techniques for calculating the class center: the first calculates the average of all data points within the class, offering a simple yet effective representation, while the second uses the median of each feature, providing a more robust alternative that minimizes the influence of extreme values. These approaches lead to the introduction of two novel models—R$^2$VFL-A and R$^2$VFL-M—each specifically designed to enhance the resilience and adaptability of the RVFL framework.
    %The Huber weighting function requires the calculation of the class center which is calculated using two techniques; first, the average of the data points of the class and second, taking median of each feature, which results in two different models R$^2$VFL-A and R$^2$VFL-M, respectively. 
    \item The experimental evaluation and statistical tests on 47 UCI datasets, demonstrate the superiority of the proposed models. 
    \item The proposed models have an application in Electroencephalogram (EEG) signal classification, demonstrating their effectiveness in handling real-world biomedical data. 
    %The proposed models has an application in classifying the EEG signals. 
\end{itemize}

\section{Related work}
Assume set $\mathcal{S}=\{x_1,x_2,\ldots,x_l\}$ denote the input training data points with each $x_i \in \mathbb{R}^n$ and $y_i \in \mathbb{R}^m$, where $m$ represents the number of classes. Let $X$ and $Y$ denote the input and target matrices of order $l \times n$ and $l \times m$, respectively.  

\subsection{Random vector functional link (RVFL) network \cite{pao1994learning}}
The structure of the RVFL neural network consists of three main layers: input, hidden, and output layers. The biases and the weights connecting the input and hidden layers, are randomly initialized and remain fixed throughout the training process. In the RVFL network, the output layer is connected to both the hidden layer features and the original input features through direct linkages. Analytical techniques such as the least-squares method or the Moore-Penrose pseudo-inverse are employed to determine the output weights, ensuring efficient and precise computation. 

Let $\mathcal{A}_1$  represent the hidden layer matrix obtained by projecting the input matrix $X$ using randomly initialized weights and biases and applying a nonlinear activation function $\Psi$. The hidden layer matrix $\mathcal{A}_1$ is defined as:
\begin{align}
    \mathcal{A}_1=\Psi(XW_1+B_1),
\end{align}
where $W_1 \in \mathbb{R}^{n \times \mathcal{L}}$ is the matrix of randomly chosen weights connecting the input layer to the hidden layer, and $B_1$ is the bias matrix of order $l \times \mathcal{L}$. Explicitly, $\mathcal{A}_1$ is written as: 
\begin{align*}
    \mathcal{A}_1= \begin{bmatrix}
        \Psi(x_1w_1+b_1) & \cdots & \Psi(x_1 w_\mathcal{L} +b_h) \\
        \vdots & \ddots & \vdots \\
        \Psi(x_l w_1 +b_1) & \cdots & \Psi(x_l w_\mathcal{L} +b_h)\\ 
    \end{bmatrix},
\end{align*}
where $w_i \in \mathbb{R}^{n \times 1}$ and $b_i$ denote the $i^{th}$ column of weight matrix $W_1$ and bias term of $i^{th}$ hidden node. $w_i$ connects all the input layer to the $i^{th}$ hidden node. The input and hidden features are concatenated and expressed as:
\begin{align*}
    \mathcal{A}_2=\begin{bmatrix}
X  & \mathcal{A}_1
    \end{bmatrix} \in \mathbb{R}^{l \times (n+\mathcal{L})}. 
\end{align*}
Assuming $W_2 \in \mathbb{R}^{(n+\mathcal{L}) \times m}$ as the weight matrix connecting the input and hidden layer with the output layer. The RVFL model's predicted output $\Hat{Y}$ is provided as follows:
\begin{align*}
    \Hat{Y}= \mathcal{A}_2 W_2.
\end{align*}
The RVFL model's objective function with $\mathcal{L}$ hidden nodes can be expressed as follows:
\begin{align}{\label{eq:rvfl}}
    \underset{W_2}{\text{ min }} \frac{1}{2}\norm{W_2}^2+\frac{\gamma}{2} \norm{\mathcal{A}_2W_2-Y},
\end{align}
where $\gamma$ denotes the positive regularization hyperparameter. 
The solution corresponding to the optimization problem (\ref{eq:rvfl}) is given as follows: 
\begin{align}
W_2=\begin{cases}
    (\mathcal{A}_2^t \mathcal{A}_2+\frac{1}{\gamma} I)^{-1} \mathcal{A}_2^tY, & (n+\mathcal{L}) \leq  l, \\
    \mathcal{A}_2^t(\mathcal{A}_2 \mathcal{A}_2^t +\frac{1}{\gamma} I)^{-1} Y, & l < n+\mathcal{L}.
\end{cases}
\end{align}

\subsection{Class probability \cite{kumari2024class}}
To reduce the impact of noise during model training, each training data point is assigned a class probability value. This value is determined by calculating the proportion of samples from the same class within a specified neighborhood to the total number of data points in that neighborhood. By accounting for local class distributions, this method enhances the stability of the training process and minimizes the effects of noise. The class probability for the $i^{th}$ data point is denoted by $cp_i$ and expressed as follows:
\begin{align}
    cp_i=\frac{|x_i: \norm{\theta(x_i)-\theta(x_j)} \leq \Delta, y_i=y_j|}{|x_i: \norm{\theta(x_i)-\theta(x_j)} \leq \Delta|},
\end{align}
where $\Delta$ is a small positive number. 
\section{Proposed work}
% Despite RVFL being an effective model for classification. The assumption of the equal contribution of all the data points in the training process affects the performance of the model \cite{mishra2023intuitionistic} in the presence of noise and outliers. To handle this issue, we propose RVFL with a Huber weighting scheme and
% class probability, called R$^2$VFL. The proposed R$^2$VFL utilizes a Huber weighting function and class probability scheme to handle the noise and outliers. 
Although RVFL is an effective model for classification, its performance can be adversely affected by the assumption that all data points contribute equally during the training process \cite{malik2022alzheimer}, particularly in the presence of noise and outliers. To address this limitation, we propose an enhanced model, R$^2$VFL (RVFL with Huber weighting scheme and class probability). The R$^2$VFL model incorporates a Huber weighting function \cite{schuster2011robust} and a class probability scheme \cite{kumari2024bell}, effectively mitigating the impact of noise and outliers while improving robustness and reliability.

\subsection{Contribution score}
Each training data point is assigned a weight based on its significance in the training process, with noise and outliers receiving lower scores to minimize their influence. The weighting scheme combines an updated Huber weighting function with class probability to achieve this.\\
     \textbf{Improved Huber weighting function: } The weights are determined by comparing the distance of each data point from its class center to a predefined threshold $\tau$. If the distance $d_{ji}$ of $i^{th}$ data point of $j^{th}$ class from its own class center is less than or equal to $\tau$, the weight assigned is 1. Otherwise, the weight is calculated as the ratio of the threshold $\tau$ to the distance $d_{ji}$. This ensures a smooth adjustment of weights, reducing the impact of distant data points (potential outliers). Mathematically, the weights $m_i$ is expressed as:
 
 % if the distance of the data points with the class center is less than or equal to the threshold value $\tau_1$, then weight assigned is 1, else the weight assigned to the data points is the ratio between threshold distance $\tau_1$ to that of the distance between the class centre and data point $d_{ji}$. Mathematically, the weights $m_i$ is defined as:
    \begin{align}
        m_i= \begin{cases}
            1, & d_{ji} \leq \tau, \\
            \frac{\tau}{d_{ji}}, & d_{ji} > \tau,  
        \end{cases}
    \end{align}
where $j=1,2,\ldots,m$ corresponds to the distinct classes; $i=1,2,\ldots,l$ corresponds to the various data points. The distance $d_{ji}$ is calculated as: 
\begin{align}
    d_{ji}=\norm{\theta(x_i)-\mathcal{C}_j},
\end{align}
where $\theta$ denotes mapping to higher dimensional space and $\mathcal{C}_j$ indicate the center of the $j^{th}$ class which is solved as shown in \cite{ganaie2024eeg}. \\

\textbf{Calculation of class center:} The class center for each class is calculated in two different ways:  
\begin{itemize}
    \item \textbf{Class averaging scheme:} The class averaging scheme provides an efficient and straightforward method for determining class centers by calculating the mean of all data points within a class. This approach ensures computational simplicity, equal treatment of data points, and accurate representation of the class distribution, making it particularly effective for datasets with minimal noise or outliers. Its ease of implementation and reliability make it a practical choice for various tasks. It is calculated as:
    \begin{align*}
        \mathcal{C}_j=\frac{1}{l_j} \sum_{i=1}^{l_j} \theta(x_j),
    \end{align*}
    where $l_j$ represents the number of samples in the $j^{th}$ class. 

    \item \textbf{Median-based centering scheme: } The median-based centering scheme identifies the class center by calculating the median of each feature within a class, offering a robust alternative to mean-based methods. Unlike averaging, it resists the influence of outliers and noise, ensuring a reliable and stable representation of the class center. This approach is particularly effective for datasets with skewed distributions or extreme values, capturing the true central tendency while maintaining robustness in diverse data scenarios.\\
To calculate the center of each class, we first project the data points of each class using the kernel function $K$. (Suppose $\mathcal{G}$ denotes samples of a class, then $K(\mathcal{G}, \mathcal{G})$ denotes the projected samples.) After projection, the data points corresponding to the $j^{th}$
  class are denoted by $x_{k}^h$, where $k=1,2,\ldots,l_j$ and $j=1,2,\ldots,m$ represents the samples in class $j$, with $l_j$  being the total number of samples in that class.

Mathematically, for the $j^{th}$ class, the class center $\mathcal{C}_j$ is computed as:
 \begin{align*}
        \mathcal{C}_j= [\text{median}(x_{k1}^h, x_{k2}^h,\ldots,x_{kn}^h)]_{k=1}^{l_j} \text{ for } j=1,2,\ldots,m, 
    \end{align*}
where $x_{ki}^h$  represents the $i^{th}$  feature of the $k^{th}$ sample in the projected space, and the median is calculated for each feature across all $l_j$  samples of the $j^{th}$ class. This ensures that the class center is resistant to outliers and reflects the true central tendency of the data.
    % To calculate the center of each class, we projected the data points of its class using the kernel function. After projection, let the data points be denoted by $x_{jk}^h$ for $j^{th}$ class samples; $j,k=1,2, \ldots,l_j$.     
    
    % Mathematically, for $j^{th}$ class, center is defined as: 
    % \begin{align*}
    %     \mathcal{C}_j= [\text{median}(x_{1k}^h, x_{2k}^h,\ldots,x_{l_jk}^h)]_{k=1}^{l_j}
    % \end{align*}
\end{itemize}
% Combining the improved weighing function with that of class probability for each of the data points, we assign the contribution score, denoted as $R_i$ to each of the data points. Empirically, $C_i$ is defined as:
% \begin{align}
%     R_i=cp_i \times m_i. 
% \end{align}
% The weights of each of the sample is multiplied with the weights to ensure their actual contribution. Thus, with their weights, 
% \begin{align}
%     \mathcal{B}_2= \mathcal{R} \mathcal{A}_2,
% \end{align}
% where $\mathcal{R}=\text{Diag}\{R_i\}$, for $i=1,2,\ldots,l$. After calculating the weights, the optimization problem of proposed R$^2$VFL is given by:
% \begin{align}
%     \underset{W_2}{\text{ min }} \norm{W_2}^2+\frac{\gamma}{2} \norm{\mathcal{B}_2W_2-\mathcal{R} Y}
% \end{align}

\subsection{Huber and class probability weighted RVFL (R$^2$VFL)}    
By integrating the improved Huber weighting function with class probability for each data point, we assign a contribution score, $R_i$, to each sample. This score is computed as the product of the class probability $cp_i$ and the corresponding Huber weight $m_i$, represented mathematically as:
\begin{align}
    R_i=cp_i \times m_i. 
\end{align}
This ensures that each data point’s contribution to the training process is scaled according to its importance, where the impact of noisy or outlying samples is reduced. To reflect these contributions, we construct the matrix $\mathcal{B}_2$ as follows:
 \begin{align}
    \mathcal{B}_2= \mathcal{R} \mathcal{A}_2,
\end{align} 
where $\mathcal{R}$ is a diagonal matrix with entries $R_i$ for $i=1,2,\ldots,l$. With these score values calculated, the optimization problem for the proposed R$^2$VFL model becomes: 
\begin{align}\label{eq:proposed}
    \underset{W_2}{\text{ min }} \norm{W_2}^2+\frac{\gamma}{2} \norm{\mathcal{B}_2W_2-\mathcal{R} Y}.
\end{align} 
This formulation balances the model’s regularization with the adjusted contributions from each sample, ensuring that noise and outliers have a minimal effect on the learning process while maintaining the stability and performance of the model. 
% Assume
% \begin{align}{\label{eq:proposedlag}}
%     \Gamma=\norm{W_2}^2+\frac{\gamma}{2} \norm{\mathcal{B}_2W_2-\mathcal{R} Y}.
% \end{align}
% To obtain the optimal $W_2$, differentiate the equation (\ref{eq:proposedlag}) with respect to $W_2$ and equate to 0. 
% \begin{align}
%     \frac{\partial \Gamma}{\partial W_2}= W_2+\gamma \mathcal{B}_2^t (\mathcal{B}_2W_2-\mathcal{R}Y)=0.
% \end{align}
On solving the optimization problem (\ref{eq:proposed}), the weights $W_2$ is obtained as follows: 
\begin{align}
    W_2=\begin{cases}
        \left( \frac{I}{\gamma}+\mathcal{B}_2^t\mathcal{B}_2\right)^{-1} \mathcal{B}_2^t \mathcal{R} Y, & (n+\mathcal{L}) \leq l, \\
        \mathcal{B}^t\left( \frac{I}{\gamma}+\mathcal{B} \mathcal{B}^t\right)^{-1} \mathcal{R}Y, & (n+\mathcal{L}) > l.
    \end{cases}
\end{align}
 Based on two distinct approaches for calculating class centers, we propose two models, R$^2$VFL-A and R$^2$VFL-M. The proposed R$^2$VFL-A employs the class averaging method for center calculation, while R$^2$VFL-M utilizes a median-based centering approach.    

\section{Numerical expermients}
To examine the performance of the proposed R$^2$VFL-A and R$^2$VFL-M models, we conducted extensive experiments which include RVFL \cite{pao1994learning}, RVFLwoDL (ELM) \cite{huang2006extreme}, minimum class variance extreme learning machine  (MCVELM) \cite{iosifidis2013minimum},  Total variance minimization-based RVFL (Total-var-RVFL) \cite{ganaie2020minimum}, Intuitionistic fuzzy RVFL (IFRVFL) \cite{malik2022alzheimer}, and neuro-fuzzy RVFL (NF-RVFL) \cite{sajid2024neuro}.

\begin{table*}[h!]
 \centering
 \caption{\vspace{0.1mm} Performance of the proposed R$^2$VFL-A, R$^2$VFL-M and the baseline models on binary UCI datasets.}
 \vspace{0.05cm}
 \resizebox{18.8cm}{!}{
\label{tab:real-world classification data}
\begin{tabular}{|lcccccccccc|}
\hline
 Dataset &RVFL \cite{pao1994learning}& RVFLwoDL \cite{huang2006extreme}&	Total-Var-RVFL \cite{ganaie2020minimum} & MCVELM \cite{iosifidis2013minimum}&	IFRVFL \cite{malik2022alzheimer} & NF-RVFL-K \cite{sajid2024neuro} &	NF-RVFL-C \cite{sajid2024neuro} &	NF-RVFL-R \cite{sajid2024neuro}&	Proposed R$^2$VFL-A &	Proposed R$^2$VFL-M \\
&Accuracy & Accuracy & Accuracy & Accuracy & Accuracy & Accuracy & Accuracy& Accuracy & Accuracy & Accuracy \\
 \hline 
bank&$89.4051$&$89.4934$&$89.6705$&$89.6703$&$89.1173$&$89.4492$&$89.9139$&$89.4269$&$89.6042$&$89.6705$\\
blood&$76.5065$&$76.9056$&$76.9065$&$77.3065$&$77.4398$&$77.8398$&$77.4398$&$77.838$&$80.8824$&$79.2781$\\
breast\_cancer\_wisc&$88.5653$&$87.9897$&$88.704$&$88.8448$&$89.8499$&$88.4183$&$87.705$&$88.2785$&$97.4278$&$97.4269$\\
breast\_cancer\_wisc\_diag&$93.8503$&$92.2683$&$94.1997$&$93.4995$&$89.2719$&$94.3782$&$95.6063$&$94.3751$&$97.3678$&$97.3678$\\
breast\_cancer\_wisc\_prog&$81.359$&$80.3846$&$82.359$&$81.8718$&$78.359$&$81.8846$&$80.9103$&$83.359$&$81.8776$&$81.8776$\\
conn\_bench\_sonar\_mines\_rocks&$62.079$&$60.5226$&$64.4251$&$64.4251$&$54.8316$&$66.2602$&$63.4611$&$65.8885$&$78.8462$&$77.8846$\\
credit\_approval&$85.2174$&$85.3623$&$85.5072$&$85.5072$&$86.5217$&$85.942$&$85.7971$&$85.942$&$86.9556$&$87.2446$\\
cylinder\_bands&$66.4154$&$65.8081$&$67.7727$&$66.0156$&$63.4875$&$69.724$&$70.1294$&$68.9606$&$68.9453$&$70.5078$\\
echocardiogram&$84.6724$&$83.9031$&$85.4701$&$85.4986$&$80.7692$&$85.4416$&$85.4416$&$85.4416$&$86.3163$&$87.0739$\\
haberman\_survival&$73.4902$&$73.8181$&$74.4738$&$73.4902$&$75.1348$&$75.1296$&$75.4574$&$75.1296$&$77.4351$&$77.1147$\\
hepatitis&$85.1613$&$82.5806$&$85.8065$&$85.1613$&$85.8065$&$87.0968$&$85.8065$&$85.8065$&$86.4879$&$87.753$\\
ilpd\_indian\_liver&$71.5311$&$72.2149$&$72.5626$&$72.5508$&$72.7277$&$72.3917$&$72.2178$&$72.3932$&$73.9266$&$73.5924$\\
ionosphere&$88.6358$&$86.9175$&$89.4809$&$88.326$&$84.3581$&$89.7626$&$88.9095$&$88.3461$&$88.6298$&$88.9139$\\
mammographic&$79.9196$&$79.2978$&$80.0237$&$80.2332$&$79.7107$&$79.504$&$79.1899$&$79.2957$&$83.2516$&$82.8328$\\
molec\_biol\_promoter&$72.7706$&$74.632$&$78.2684$&$72.684$&$78.3983$&$82.0779$&$82.0346$&$82.0346$&$79.2735$&$81.2322$\\
musk\_1&$72.0614$&$69.7675$&$72.4846$&$70.1776$&$71.864$&$76.0482$&$74.1645$&$75.2149$&$71.8487$&$74.1597$\\
oocytes\_merluccius\_nucleus\_4d&$82.2884$&$81.5045$&$82.3883$&$82.1942$&$79.1554$&$81.7006$&$82.1923$&$81.6021$&$83.7596$&$83.8576$\\
ozone&$97.1217$&$97.1217$&$97.1611$&$97.1217$&$97.1612$&$97.1612$&$97.1612$&$97.2005$&$97.1609$&$97.1609$\\
parkinsons&$80.5128$&$80.5128$&$82.0513$&$83.5897$&$78.4615$&$83.0769$&$84.1026$&$83.0769$&$84.003$&$86.5434$\\
pima&$72.0075$&$72.2698$&$72.92$&$72.9166$&$73.8282$&$72.9174$&$72.6594$&$73.6983$&$78.125$&$78.125$\\
pittsburg\_bridges\_T\_OR\_D&$87.1905$&$87.1905$&$89.1429$&$90.1429$&$89.1905$&$88.2381$&$90.1905$&$90.1905$&$87.1923$&$88.1539$\\
planning&$71.3814$&$71.3814$&$72.4925$&$73.048$&$69.8048$&$71.9369$&$71.9369$&$73.018$&$73.0435$&$72.4879$\\
spect&$68.3019$&$66.7925$&$69.0566$&$67.9245$&$68.3019$&$69.434$&$68.6792$&$69.0566$&$74.7343$&$73.2248$\\
spectf&$79.3431$&$79.7205$&$79.7135$&$79.7205$&$79.3431$&$79.7205$&$79.7205$&$79.7205$&$85.7361$&$85.7418$\\
statlog\_australian\_credit&$68.2609$&$68.4058$&$68.6957$&$68.5507$&$68.9855$&$68.8406$&$68.8406$&$69.2754$&$68.4114$&$68.8441$\\
statlog\_german\_credit&$76.9$&$75.7$&$77$&$76.4$&$76.7$&$77.7$&$77.7$&$77.1$&$76.6$&$76.7$\\
statlog\_heart&$80.3704$&$80$&$81.4815$&$82.5926$&$81.8519$&$82.2222$&$81.8519$&$81.4815$&$86.2928$&$85.9361$\\
tic\_tac\_toe&$88.8264$&$88.9278$&$91.8521$&$90.0769$&$81.4316$&$68.5684$&$79.3216$&$65.3125$&$98.5417$&$100$\\
titanic&$77.9168$&$77.9168$&$78.6901$&$79.0532$&$79.0532$&$79.0532$&$79.0532$&$78.4623$&$80.0976$&$82.3702$\\
vertebral\_column\_2clases&$71.2903$&$70.6452$&$72.5806$&$74.5161$&$85.4839$&$81.6129$&$80.3226$&$79.0323$&$85.4854$&$85.4854$\\
\hline
Average Accuracy &  79.1118	&78.6652&	80.1114	&79.7703&	78.88&	80.1177	&80.2639&	79.8653&	82.942&	83.2854\\	
% Average Rank &8.5 &	9.33&	8.6	&9.3&	5.57&	6.47&	6.93&	4.67&	5	&5.17&	3.37&	2.53\\
Average Rank &8	&8.73 &	5.27&	6.12	&6.62&	4.48	&5.05&	5&	3.22&	2.52\\
\hline 
\end{tabular}}
% \footnotesize{{ Bold values represent the best G-mean values.}}
\end{table*}

% \subsection{Experimental Setup}
\subsection{Experimental setup}
The following section outlines the parameter selection ranges for the proposed R$^2$VFL models and the baseline models used for comparison. The experiments are conducted on a computing system featuring MATLAB R2023a, an  Intel(R)Xeon(R) Platinum 8260 CPU @ 2.30GHz, with 256 GB RAM on a Windows-10 operating platform. We followed the experimental setup outlined in \cite{sajid2024neuro}. A 5-fold cross-validation strategy combined with grid search is implemented for hyperparameter tuning. The testing accuracy is computed independently for each fold for every set of hyperparameters, and the average accuracy across all five folds is calculated. The optimal parameters leading to the highest average testing accuracy are selected.  The regularization parameter $\gamma$ for the proposed models, R$^2$VFL-A and R$^2$VFL-M, and the baseline models are chosen from range $[10^{-5},10^{-4},\ldots,10^5]$; the number of hidden nodes  $\mathcal{L}$ is selected from the range $(3:20:203)$. For the proposed NF-RVFL model, the number of fuzzy rules in the fuzzy layer is selected from the range $(5:5:50)$ with the standard deviation set to 1 \cite{sajid2024neuro}. For the IFRVFL model, the kernel parameter is chosen from the set $\{2^{-5},2^{-4},\ldots,2^5\}$. For the GEELMLDA, GEELM-LFDA, Total-Var-RVFL, and MCVELM models, 
another hyperparameter is varied within the set $\{10^{-5},10^{-4},\ldots,10^5\}$. The predefined threshold parameter $\tau$ takes the range [1/2, 5/8, 3/4, 7/8, 1]$\times$(radius of
positive/negative class). The RBF kernel is used for the projection of the samples. 

\begin{table*}[h!]
\centering
\caption{Nemenyi post-hoc analysis demonstrating significant variations between the proposed R$^2$VFL-M and the baseline models over binary UCI datasets.}
\resizebox{.7\textwidth}{!}{% <------ Don't forget this %
  \begin{tabular}{|lcccccccc|}
    \hline 
   &RVFL \cite{pao1994learning}& RVFLwoDL \cite{huang2006extreme}&	Total-Var-RVFL \cite{ganaie2020minimum} & 	MCVELM \cite{iosifidis2013minimum}	& IFRVFL \cite{malik2022alzheimer} & NF-RVFL-K \cite{sajid2024neuro} &	NF-RVFL-C \cite{sajid2024neuro} &	NF-RVFL-R \cite{sajid2024neuro}\\
    \hline
   Proposed R$^2$VFL-M & Yes  &  Yes & Yes &  Yes & Yes & No & Yes  & Yes \\
   % Proposed SRLS-OCSVM & No & Yes  & Yes & Yes\\
         \hline
     \end{tabular}
     \label{tab:Nemenyiposthoc_UCI}
}
\end{table*}

\begin{table*}[h!]
\centering
\caption{Comparison of the baseline models with the proposed R$^2$VFL-A and R$^2$VFL-M across binary UCI datasets using the Wilcoxon signed-rank test.}
\resizebox{.55\textwidth}{!}{% <------ Don't forget this %
   \begin{tabular}{|l|cccc|cccc|}
    \hline 
    & \multicolumn{4}{|c|}{Proposed R$^2$VFL-A} & \multicolumn{4}{|c|}{Proposed R$^2$VFL-M}\\
    \hline
    Model & $R^+$ & $R^-$ & $p$-value & Null hypothesis& $R^+$ & $R^-$ & $p$-value & Null hypothesis\\
    \hline
  RVFL \cite{pao1994learning} & $450$ & $15$ & $< 0.00001$ & rejected & $463$ & $2$ & $< 0.00001$ & rejected  \\
   RVFLwoDL \cite{huang2006extreme} & $465$ & $0$ & $<0.00001$  & rejected & $465$ & $0$ & $<0.00001$ & rejected\\ 
   % GEELM-LDA \cite{iosifidis2015graph} & $401$ & $34$ & $0.00008$ & rejected & $439$ & $26$ & $<0.00001$ & rejected \\ 
   % GEELM-LFDA \cite{iosifidis2015graph} & $449$ & $16$ & $<0.00001$ & rejected & $451$ & $14$ & $<0.00001$ & rejected\\
   Total-Var-RVFL \cite{ganaie2020minimum}   & $416$ & $49$ & $0.00016$ & rejected & $410$ & $25$ & $<0.00001$ & rejected\\
    MCVELM \cite{iosifidis2013minimum} & $438$ & $27$ & $<0.00001$ & rejected & $446$ & $19$ & $<0.00001$ & rejected\\ 
   IFRVFL \cite{malik2022alzheimer} & $438$ & $27$ & $<0.00001$ & rejected & $424$ & $11$ & $<0.00001$ & rejected\\
   NF-RVFL-K \cite{sajid2024neuro}  & $368$ & $97$ & $0.00528$ & rejected & $411$ & $54$ & $0.00024$ & rejected\\
   NF-RVFL-C \cite{sajid2024neuro} & $370$ & $95$ & $0.00466$ & rejected & $418$ & $47$ & $0.00014$ & rejected\\
   NF-RVFL-R \cite{sajid2024neuro} & $378$ & $87$ & $0.00278$ & rejected & $408$ & $57$ & $0.0003$ & rejected\\
         \hline
     \end{tabular}}
     \label{tab:Wilcoxon_UCI}
\end{table*}

\begin{table*}[h!]
 \centering
 \caption{\vspace{0.1mm} Performance of the proposed R$^2$VFL-A, R$^2$VFL-M and the baseline models on multiclass UCI datasets.}
 \vspace{0.05cm}
 \resizebox{18.0cm}{!}{
\label{tab:multiclass classification data}
\begin{tabular}{|lccccccccc|}
\hline
 Dataset &RVFL \cite{pao1994learning}& RVFLwoDL \cite{huang2006extreme}&	Total-Var-RVFL \cite{ganaie2020minimum} & 	MCVELM \cite{iosifidis2013minimum}	& NF-RVFL-K \cite{sajid2024neuro} &	NF-RVFL-C \cite{sajid2024neuro} &	NF-RVFL-R \cite{sajid2024neuro}&	Proposed R$^2$VFL-A &	Proposed R$^2$VFL-M \\
&Accuracy & Accuracy & Accuracy & Accuracy & Accuracy & Accuracy & Accuracy & Accuracy & Accuracy   \\
 \hline 
abalone&$63.4665$&$63.4419$&$63.754$&$63.6578$&$64.1365$&$64.0887$&$63.8253$&$65.0708$&$64.9273$\\
conn\_bench\_vowel\_deterding&$95.8586$&$95.1515$&$96.1616$&$95.7576$&$88.7879$&$92.2222$&$87.9798$&$97.8818$&$98.1843$\\
contrac&$40.458$&$40.1183$&$41.3386$&$41.0688$&$42.3673$&$41.5416$&$41.4934$&$43.9358$&$43.8021$\\
dermatology&$97.5379$&$97.0011$&$97.2714$&$97.5417$&$97.8193$&$97.5379$&$97.5454$&$97.5394$&$97.5394$\\
ecoli&$60.9175$&$61.2116$&$51.0667$&$51.0667$&$60.6277$&$61.5145$&$60.619$&$80.0595$&$80.6548$\\
energy\_y1&$88.6716$&$88.2777$&$89.7106$&$89.0604$&$88.5434$&$89.0595$&$87.889$&$89.0625$&$89.0625$\\
energy\_y2&$90.4932$&$90.4932$&$91.14$&$90.7538$&$89.1928$&$89.8404$&$89.3269$&$89.1927$&$89.1927$\\
glass&$37.6855$&$39.0808$&$34.4297$&$37.6855$&$42.8239$&$40.4873$&$42.3477$&$54.5161$&$53.992$\\
hayes\_roth&$61.875$&$60.625$&$62.5$&$61.875$&$61.875$&$65$&$61.875$&$67.5$&$65.625$\\
heart\_cleveland&$59.7268$&$59.377$&$61.0328$&$59.3607$&$61.3388$&$60.0437$&$61.0383$&$62.0175$&$62.0175$\\
heart\_va&$40$&$40$&$40.5$&$40$&$42.5$&$41.5$&$41$&$39$&$40.5$\\
led\_display&$72.6$&$72.5$&$73.5$&$72.8$&$72.2$&$73.4$&$72.7$&$73.4$&$72.7$\\
oocytes\_merluccius\_states\_2f&$91.2846$&$90.8948$&$91.8737$&$91.483$&$91.1894$&$92.067$&$91.3845$&$92.4636$&$92.6589$\\
semeion&$87.0673$&$82.4233$&$87.2562$&$83.0492$&$88.5732$&$87.3187$&$88.0726$&$90.3954$&$90.332$\\
statlog\_image&$95.1948$&$94.8918$&$95.6277$&$95.2814$&$91.5584$&$93.0736$&$91.5584$&$95.0649$&$95.1949$\\
statlog\_vehicle&$82.0327$&$81.5593$&$82.0313$&$81.3227$&$79.5475$&$82.2666$&$79.1974$&$83.3335$&$83.3335$\\
vertebral\_column\_3clases&$65.1613$&$65.1613$&$84.1935$&$84.1935$&$69.0323$&$65.4839$&$67.4194$&$73.5348$&$71.4286$\\
\hline
Average Accuracy &  72.3548 &	71.8946&	73.1405&	72.7034	&72.4773&	72.7321	&72.0748	&76.1158&	75.9497	\\	
% Average Rank &6 &	7.24&	4.29&	5.29&	5.18&	4.53&	5.71&	2.71&	2.65\\
Average Rank & 6.26	&7.35&	4.38&	5.53&	5.29&	4.59&	5.85	&2.88&	2.85\\
\hline 
\end{tabular}}
% \footnotesize{{ Bold values represent the best G-mean values.}}
\end{table*}

\subsection{Analysis of the results across binary datasets}

The experimental results conducted on 30 binary UCI datasets \cite{dua2017uci} are presented in Table \ref{tab:real-world classification data}. This table presents the average classification accuracy of the proposed and baseline models. Among the models, the proposed R$^2$VFL-M achieved the best average accuracy of 83.2854\%, followed by R$^2$VFL-A, which achieved the second-best average accuracy of 82.942\%. In comparison, the baseline models showed the following average accuracies: RVFL achieved 79.1118\%, RVFLwoDL recorded 78.6652\%, Total-Var-RVFL achieved 80.1114\%, MCVELM achieved 79.7703\%, IFRVFL achieved 78.88\%, NF-RVFL-K achieved 80.1177\%, NF-RVFL-C achieved 80.2639\%, and NF-RVFL-R achieved 79.8653\%. However, relying solely on average accuracy can be misleading, as high performance on specific datasets may disproportionately influence the overall average, potentially resulting in an unreliable measure of performance. To address this limitation, ranks are assigned to all models based on their performance across the datasets, where a better-performing model is assigned a lower rank. The average rank is then calculated for each model to provide a more robust performance evaluation. The average ranks highlight the superiority of the proposed models, with proposed R$^2$VFL-M achieving the best average rank of 2.52, and proposed R$^2$VFL-A following closely with an average rank of 3.22. Among the baseline models, the average ranks are as follows: RVFL (8), RVFLwoDL (8.73), Total-Var-RVFL (5.27), MCVELM (6.12), IFRVFL (6.62), NF-RVFL-K (4.48), NF-RVFL-C (5.05), and NF-RVFL-R (5). These results demonstrate the consistent and superior performance of the proposed models. The proposed R$^2$VFL-M emerges as the best-performing model, with the proposed R$^2$VFL-A ranking as the second-best, reaffirming their outstanding performance compared to the baseline models. \par

\begin{table*}[h!]
 \centering
 \caption{\vspace{0.1mm} Performance of the proposed R$^2$VFL-A, R$^2$VFL-M and the baseline models on EEG signal classification.}
 \vspace{0.05cm}
 \resizebox{17.7cm}{!}{
\label{tab:EEG classification data}
\begin{tabular}{|lccccccc|}
\hline
 Dataset &RVFL \cite{pao1994learning}& RVFLwoDL \cite{huang2006extreme}&	Total-Var-RVFL \cite{ganaie2020minimum} & 	MCVELM \cite{iosifidis2013minimum}& IFRVFL \cite{malik2022alzheimer}	& 	Proposed R$^2$VFL-A &	Proposed R$^2$VFL-M \\
&Accuracy & Accuracy & Accuracy & Accuracy & Accuracy &  Accuracy & Accuracy   \\
 \hline 
 EEG\_o\_vs\_f\_entropy\_50&$67.5$&$69.5$&$71$&$72$&$68$&$75$&$75$\\
EEG\_o\_vs\_f\_ttest\_50&$76$&$76$&$79$&$78.5$&$78.5$&$81.5$&$81$\\
EEG\_o\_vs\_n\_ttest\_50&$81.5$&$79$&$81.5$&$81.5$&$80.5$&$81$&$82.5$\\
EEG\_o\_vs\_n\_wilcoxon\_50&$82.5$&$84$&$84$&$83.5$&$83.5$&$84$&$84$\\
EEG\_z\_vs\_f\_roc\_50&$82.5$&$79.5$&$83.5$&$83$&$82$&$85$&$84.5$\\
EEG\_z\_vs\_f\_wilcoxon\_50&$80.5$&$80$&$81.5$&$81.5$&$80$&$81.5$&$81$\\
EEG\_z\_vs\_n\_wilcoxon\_50&$83.5$&$81$&$83$&$83.5$&$82.5$&$83$&$83.5$\\
EEG\_o\_vs\_n\_wilcoxon\_100&$87$&$86.5$&$87.5$&$87$&$88.5$&$88.5$&$89.5$\\
EEG\_z\_vs\_f\_roc\_100&$86$&$84.5$&$87$&$86.5$&$87.5$&$87.5$&$87.5$\\
EEG\_z\_vs\_f\_ttest\_100&$81$&$80$&$80.5$&$81$&$80.5$&$81.5$&$81.5$\\
EEG\_z\_vs\_f\_wilcoxon\_100&$85.5$&$85$&$87$&$85.5$&$85$&$89$&$88.5$\\
EEG\_z\_vs\_n\_bhattacharyya\_100&$76.5$&$76.5$&$77.5$&$77.5$&$75.5$&$77.5$&$77.5$\\
EEG\_f\_vs\_s\_ttest\_150&$92$&$92$&$92$&$93$&$93$&$93$&$93$\\
EEG\_o\_vs\_n\_roc\_150&$79.5$&$78.5$&$80.5$&$80.5$&$78.5$&$81$&$82.5$\\
EEG\_o\_vs\_n\_wilcoxon\_150&$88$&$86$&$88.5$&$88.5$&$88$&$88$&$88.5$\\
EEG\_z\_vs\_f\_ttest\_150&$78.5$&$80$&$80$&$81.5$&$78.5$&$79$&$80.5$\\
EEG\_z\_vs\_n\_entropy\_150&$76$&$76$&$77$&$76.5$&$76.5$&$76$&$77$\\
EEG\_z\_vs\_n\_roc\_150&$82$&$82$&$82$&$82$&$80.5$&$80.5$&$82.5$\\
EEG\_z\_vs\_n\_ttest\_150&$73.5$&$71.5$&$74$&$74.5$&$73$&$75$&$75$\\
EEG\_z\_vs\_n\_wilcoxon\_150&$86.5$&$86.5$&$87.5$&$88$&$86.5$&$88$&$87.5$\\
EEG\_f\_vs\_s\_roc\_200&$91.5$&$91.5$&$90.5$&$91.5$&$91$&$91.5$&$91.5$\\
EEG\_n\_vs\_s\_wilcoxon\_200&$92$&$96.5$&$95$&$97$&$94$&$97$&$97$\\
EEG\_o\_vs\_f\_entropy\_200&$75.5$&$74.5$&$75$&$75$&$74$&$76$&$76$\\
EEG\_o\_vs\_f\_roc\_200&$81$&$80$&$81.5$&$81.5$&$82$&$82$&$82$\\
EEG\_o\_vs\_f\_ttest\_200&$78.5$&$79$&$80.5$&$81$&$79$&$79.5$&$80.5$\\
EEG\_o\_vs\_n\_bhattacharyya\_200&$74$&$74$&$75.5$&$77.5$&$75$&$76$&$76$\\
EEG\_o\_vs\_n\_wilcoxon\_200&$88.5$&$88.5$&$89.5$&$89$&$89.5$&$90$&$90$\\
EEG\_z\_vs\_f\_bhattacharyya\_200.&$72$&$73$&$75.5$&$75$&$75.5$&$80.5$&$80.5$\\
EEG\_z\_vs\_f\_entropy\_200&$72.5$&$72$&$74$&$73.5$&$73.5$&$75$&$75$\\
EEG\_z\_vs\_f\_ttest\_200.&$80.5$&$80.5$&$82$&$81.5$&$82.5$&$82.5$&$82.5$\\
EEG\_z\_vs\_f\_wilcoxon\_200&$85.5$&$86$&$87$&$86$&$85$&$86.5$&$87.5$\\
EEG\_z\_vs\_n\_bhattacharyya\_200&$77$&$76.5$&$78.5$&$78$&$77$&$79.5$&$80$\\
EEG\_z\_vs\_n\_entropy\_200&$77.5$&$76$&$78$&$78$&$79$&$81$&$80.5$\\
EEG\_z\_vs\_n\_roc\_200&$85$&$84.5$&$87.5$&$86$&$88$&$88.5$&$88.5$\\
\hline
Average Accuracy &  81.0882	&80.7794	&82.1912&	82.2353&	81.5147&	82.9706	& 83.2353\\	
% Average Rank &6 &	7.24&	4.29&	5.29&	5.18&	4.53&	5.71&	2.71&	2.65\\
Average Rank &5.46	&5.99&	3.63&	3.5	&5&	2.53&	1.9\\
\hline 
\end{tabular}}
% \footnotesize{{ Bold values represent the best G-mean values.}}
\end{table*}

\begin{table*}[h!]
\centering
\caption{Nemenyi post-hoc analysis demonstrating significant variations between the proposed R$^2$VFL-M and the baseline models over EEG datasets.}
\resizebox{.6\textwidth}{!}{% <------ Don't forget this %
  \begin{tabular}{|lccccc|}
    \hline 
   &RVFL \cite{pao1994learning}& RVFLwoDL \cite{huang2006extreme}&	Total-Var-RVFL \cite{ganaie2020minimum} & 	MCVELM \cite{iosifidis2013minimum}	& IFRVFL \cite{malik2022alzheimer} \\
    \hline
   Proposed R$^2$VFL-M & Yes  & Yes & Yes & Yes &  Yes  \\
   % Proposed SRLS-OCSVM & No & Yes  & Yes & Yes\\
         \hline
     \end{tabular}
     \label{tab:Nemenyiposthoc_EEG}
}
\end{table*}

% To further strengthen our results, we also employed statistical tests such as the Friedman test, Nemenyi-post-hoc test, and Wilcoxon signed rank test \cite{demvsar2006statistical}. The Friedman statistics $(\chi_F^2)$ is calculated as follows: $\chi_F^2=\frac{12 \mathscr{D}}{p(p+1)}[\sum_k{R_k^2}-\frac{p(p+1)^2}{4}]$, with degree of freedom $(p-1)$, where $\mathscr{D}$ and $p$ denote number of datasets and models, respectively. To address the overly conservative nature of the Friedman statistic, the $F_F$ statistic is computed as follows: $F_F=\frac{(\mathscr{D}-1) \chi_F^2}{\mathscr{D}(p-1)-\chi_F^2}$. $F_F$ statistic follows $F$ distribution having $(p-1,(p-1)(\mathscr{D}-1))$ degrees of freedom. We performed two pairwise tests, which include the Nemenyi-post-hoc test and the Wilcoxon test. The Nemenyi-post-hoc test calculates the critical difference: $c=q_{\alpha} \sqrt{\frac{p(p+1)}{6\mathscr{D}}}$. If the rank difference between the proposed model and the baseline model is greater than $c$, the respective baseline model is statistically different from the proposed model. 
% Further, the pairwise Wilcoxon test evaluates the significance of the proposed models in comparison to the baseline models. This test operates under the null hypothesis, which assumes the equivalence of the two models being compared. It ranks the differences in performance between the two models based on the magnitude of these differences. Let $R^+$ and $R^-$ represent the sum of positive and negative ranks, respectively, derived from the paired comparisons. 
We employed statistical tests to further validate our results, including the Friedman test, Nemenyi post-hoc test, and Wilcoxon signed-rank test \cite{demvsar2006statistical}. The Friedman test's null hypothesis makes the assumption that the models are equivalent. The Friedman statistics $(\chi_F^2)$ is calculated as follows: $\chi_F^2=\frac{12 \mathscr{D}}{p(p+1)}[\sum_k{r_k^2}-\frac{p(p+1)^2}{4}]$, with degree of freedom $(p-1)$, where $\mathscr{D}$ and $p$ denote number of datasets and models, respectively; $r_k$ signify the average rank of $k^{th}$ model. The $F_F$ statistic is computed to counteract the Friedman statistic's conservative tendency, using the formula: $F_F=\frac{(\mathscr{D}-1) \chi_F^2}{\mathscr{D}(p-1)-\chi_F^2}$. $F_F$ statistic follows $F$ distribution having $(p-1,(p-1)(\mathscr{D}-1))$ degrees of freedom. We also conducted two pairwise tests: the Nemenyi post-hoc test and the Wilcoxon signed-rank test. The Nemenyi test evaluates the significance of differences in models' performance by calculating the critical difference $(c)$ as: $c=q_{\alpha} \sqrt{\frac{p(p+1)}{6\mathscr{D}}}$. If the rank difference between the proposed and baseline models exceeds $c$, the baseline model is considered statistically different from the proposed model. The Wilcoxon signed-rank test further assesses the significance of the proposed models compared to the baseline models. Operating under the null hypothesis, which assumes equivalence between the two models, the test ranks the performance differences based on their magnitude. The sums of positive and negative ranks are denoted as $R^+$ and $R^-$, respectively, and these values are used to evaluate statistical significance. \par 
For binary case, we have $\mathscr{D}=30$ and $p=10$, which results in $\chi_f^2= 111.4570$  $F_F=20.3872$. The degree of freedom for $\chi_F^2$ and $F_F$ is 9 and $(9,261)$, respectively. $F_{(9,261)}=1.9158 < F_F$, thus the null hypothesis is rejected, hence the models are not equivalent. Further, using the Nemenyi post-hoc test, we obtain $c =  2.4734$ for $\alpha=0.05$ having $q_{\alpha}=3.164$. The results of the Nemenyi post-hoc test are presented in Table \ref{tab:Nemenyiposthoc_UCI}. The results for the Wilcoxon signed rank test are presented in Table \ref{tab:Wilcoxon_UCI}, which highlights the null hypothesis assumption is rejected in comparison to all baseline models.  This confirms the statistical significance between the proposed models and the baseline models. Consequently, the numerical results and statistical analysis conducted across the binary datasets demonstrate the superior performance of the proposed models.

\subsection{Analysis of the results across multiclass datasets} 
Table \ref{tab:multiclass classification data} presents the results corresponding to the 17 multiclass UCI datasets along with the average accuracies and average ranks of different models. We excluded 
IF-RVFL from multiclass comparisons as it is designed solely for binary classification.  The average accuracies of the proposed R$^2$VFL-M and R$^2$VFL-A are $75.9497\%$ and $76.1158\%$, respectively, which are the highest as compared to the baseline models having average accuracies: 72.3548\%, 71.8946\%, 73.1405\%, 72.7034\%, 72.4773\%, 72.7321\%, and 72.0748\%, for RVFL, RVFLwoDL, Total-Var-RVFL, MCVELM, NF-RVFL-K, NF-RVFL-C, and NF-RVFL-R, respectively. The average ranks of the proposed models, R$^2$VFL-A and R$^2$VFL-M, are 2.71 and 2.65, respectively, indicating their strong performance. In comparison, the average ranks of the baseline models, listed in increasing order, are as follows: Total-Var-RVFL (4.29), NF-RVFL-C (4.53), NF-RVFL-K (5.18), MCVELM (5.29), NF-RVFL-R (5.71), RVFL (6.00), and RVFLwoDL (7.24).  \par
The statistical tests are employed for multiclass datasets with $\mathcal{D}=17$ and $p=9$ which results in $\chi_F^2=40.0452$ and $F_F=6.6773$ having degree of freedom $8$ and $(8,128)$, respectively. From table, $F_{(8,128)}=2.0114$ which is less than $F_F$, hence the null hypothesis is rejected and modes under comparison are not equivalent.  Therefore, the average accuracy and average ranking emphasize the superior performance of the proposed models across multiclass datasets. Further, the statistical test affirms the non-equivalence of the proposed models with the baseline models.

\section{Application}
The proposed models R$^2$VFL-A and R$^2$VFL-M along with baseline models are applied for classifying the EEG signals. We utilized publicly available EEG data \cite{andrzejak2001indications}, which includes five sets: S, N, F, Z, and O. Each set contains 100 single-channel EEG signals sampled at 173.61Hz for a duration of 23.6 seconds. The Z and O sets represent EEG signals from subjects with eyes open and closed, respectively. The N and F sets correspond to subjects in an interictal state, while the S set contains ictal state signals from seizure activity. For feature extraction, we applied the RankFeatures() function in MATLAB using criteria such as T-Test, Entropy, Bhattacharyya, ROC, and Wilcoxon.  \par
The experimental results, presented in Table \ref{tab:EEG classification data}, demonstrate the superior performance of the proposed models. The proposed R$^2$VFL-M model achieves the highest average accuracy of 83.24\%, with the smallest average rank of 1.9. The proposed R$^2$VFL-A model follows closely with an average rank of 2.53  and an accuracy of  82.97\%. In comparison, the baseline models exhibit the following average accuracies: Total-Var-RVFL at 82.19\%, MCVELM at 82.24\%, IFRVFL at 81.51\%, RVFL at 81.09\%, ELM at 80.78\%, NF-RVFL-K at 74.53\%, NF-RVFL-R at 69.32\%, and NF-RVFL-C at 59.26\%. These results underscore the superior performance of the proposed models in EEG signal classification.  \par
To reinforce our analysis, we conducted a statistical test with $\mathcal{D}=34$ and $p=7$, resulting in $\chi_F^2=102.9435$, $F_F=33.6162$, with degrees of freedom $6$ and $(6,198)$, respectively. Since $F_{(6,198)}=2.1445<F_F$, the null hypothesis is rejected, indicating that the models are not equivalent. Furthermore, the critical difference of the proposed and baseline models with $c$ calculated as $1.5451$ is summarized in Table \ref{tab:Nemenyiposthoc_EEG}. The results indicate that the proposed R$^2$VFL-M model is statistically distinct from all the baseline models. Further, the average rank and average accuracy of the proposed models confirm their overall superiority compared to all the baseline models. Consequently, the proposed models demonstrate superior performance in classifying EEG signals.

% To strengthen our analysis, we conduct the statistical test. For this case, we have $\mathcal{D}=34$ and $p=10$, which results into  $\chi_F^2=246.1509$, $F_F=135.7242$, 
% with degrees of freedom $9$ and $(9,297)$, respectively. $F_{(9,297)}=1.911<F_F$, thus null hypothesis assumption is rejected and the models are not equivalent. Further, the critical difference calculated is $c=2.3234$, whose results are presented in Table \ref{tab:Nemenyiposthoc_EEG}, highlighting the baseline models which are statistically different from the proposed R$^2$VFL-M. 

% The experimental results are presented in Table \ref{tab:EEG classification data}, which showcase the proposed models have superior performance having the highest average rank and smallest average accuracy of 2.53 and 82.9706\% for proposed R$^2$VFL-A and 1.9 and	83.2353\% for R$^2$VFL-M, respectively. The average accuracies of the baseline models are 82.1912	82.2353	81.5147 81.0882	80.7794 	74.5294		69.3235 59.2576 for Total-Var-RVFL, MCVELM, IFRVFL, RVFL, ELM, NF-RVFL-K, NF-RVFL-R, NF-RVFL-C, respectively. 

\section{Conclusion}
This paper presents the novel R$^2$VFL model, an advanced version of the RVFL framework that combines the Huber weighting function with class probability to assign contribution scores to data points. The Huber weighting function plays a significant role in mitigating the adverse impact of outliers by assigning a membership value of 1 to data points within a predefined threshold. Beyond this threshold, the weight adjusts based on the ratio of the threshold to the distance, offering a more refined approach to handling outliers. In addition, class probability effectively reduces the influence of noise in the training data, further enhancing the model's robustness. The computation of Huber weights relies on the calculation of the class center, which is determined using two distinct methods: first, by averaging all data points within the class, providing a simple yet effective representation; second, by employing the median of each feature, which offers a more robust alternative that minimizes the impact of extreme values. These innovative approaches give rise to two distinct models—R$^2$VFL-A and R$^2$VFL-M—tailored to improve the robustness and flexibility of the RVFL framework. Extensive experiments and rigorous statistical analysis on UCI datasets demonstrate superiority of the proposed models. Notably, the proposed models exhibit exceptional performance in EEG signal classification, further validating their practical utility. While the proposed models show significant advancements, they do not possess the feature extraction and dimensionality reduction capabilities typically found in deep learning models. As a result, a key avenue for future research will be extending the proposed models to their deep and ensemble variants, which will further enhance their capabilities and broaden their applicability.  Further, research can explore integrating the improved Huber weighting function to handle matrix input samples directly, inspired by support matrix machine \cite{kumari2024support}.

\bibliographystyle{IEEEtranN}
 \bibliography{refs}
 
\end{document}